\documentclass{bmvc2k}

\usepackage{algorithm}
\usepackage{algorithmic}
\usepackage{amsmath}
\usepackage{multirow}
\usepackage{pifont}
\usepackage{graphicx}
\usepackage{tikz}
\usepackage{subcaption}
\usepackage{amsfonts}
\usepackage{booktabs}
\usepackage{xcolor}
\usepackage{multicol}

\title{Mapping like a Skeptic:  
Probabilistic BEV Projection for Online HD Mapping}

\addauthor{Fatih Erdoğan}{https://fatih-erdogan.github.io/}{1}
\addauthor{Merve Rabia Barın}{https://mrabiabrn.github.io/}{1, 2}
\addauthor{Fatma Güney}{https://mysite.ku.edu.tr/fguney/}{1, 2}

\addinstitution{
Department of Computer Engineering, Koç University, \\
Istanbul, Turkey \\
}

\addinstitution{
KUIS AI Center \\
}

\runninghead{F. Erdogan et al.}{Mapping like a Skeptic}

\def\eg{\emph{e.g}\bmvaOneDot}

\newcommand{\tick}{\ding{51}}
\newcommand{\cross}{\ding{55}}

\newcommand{\bV}{\mathbf{V}}

\newcommand{\bB}{\mathbf{B}}

\newcommand{\bF}{\mathbf{F}}

\newcommand{\bd}{\mathbf{d}}

\newcommand{\bb}{\mathbf{b}}

\newcommand{\bmu}{\boldsymbol{\mu}}

\newcommand{\bSigma}{\boldsymbol{\Sigma}}

\newcommand{\balpha}{\boldsymbol{\alpha}}
\newcommand{\nR}{\mathbb{R}}

\newcommand{\cN}{\mathcal{N}}

\newcommand{\cC}{\mathcal{C}}

\newcommand{\figref}[1]{\Fig~\ref{#1}}

\newcommand{\algref}[1]{Algorithm~\ref{#1}}

\newcommand{\tabref}[1]{Table~\ref{#1}}

\makeatletter
\DeclareRobustCommand\onedot{\futurelet\@let@token\@onedot}
\def\@onedot{\ifx\@let@token.\else.\null\fi\xspace}
\def\eg{e.g\onedot} 
\def\ie{i.e\onedot} 
 
 \def\vs{vs\onedot}

\def\Fig{Fig\onedot}   
\makeatother

\newcommand{\xdownarrow}[1]{%
  {\left\downarrow\vbox to #1{}\right.\kern-\nulldelimiterspace}
}

\newcommand{\xuparrow}[1]{%
  {\left\uparrow\vbox to #1{}\right.\kern-\nulldelimiterspace}
}

\newcommand{\boldparagraph}[1]{\vspace{0.15cm}\noindent{\bf #1:} }

\definecolor{First}{HTML}{BDE6CD}%
\definecolor{Second}{HTML}{E2EEBC}%
\definecolor{Third}{HTML}{FFF8C5}%

\begin{document}

\maketitle

\begin{abstract}
Constructing high-definition (HD) maps from sensory input requires accurately mapping the road elements in image space to the Bird's Eye View (BEV) space. The precision of this mapping directly impacts the quality of the final vectorized HD map. Existing HD mapping approaches outsource the projection to standard mapping techniques, such as attention-based ones. However, these methods struggle with accuracy due to generalization problems, often hallucinating non-existent road elements. Our key idea is to start with a geometric mapping based on camera parameters and adapt it to the scene to extract relevant map information from camera images. To implement this, we propose a novel probabilistic projection mechanism with confidence scores to (i) refine the mapping to better align with the scene and (ii) filter out irrelevant elements that should not influence HD map generation. In addition, we improve temporal processing by using confidence scores to selectively accumulate reliable information over time. Experiments on new splits of the nuScenes and Argoverse2 datasets demonstrate improved performance over state-of-the-art approaches, indicating better generalization. The improvements are particularly pronounced on nuScenes and in the challenging long perception range. Our code and model checkpoints are available at \url{https://github.com/Fatih-Erdogan/mapping-like-skeptic}.

\end{abstract}    
\section{Introduction}
\label{sec:intro}
HD maps are crucial in self-driving to provide rich semantic information about roads. Due to the high cost of maintaining offline maps, online solutions that use onboard sensors are more desirable. 
In online HD map estimation, ensuring the accuracy of predicted road elements is critical to safe navigation. Recent approaches estimate vectorized HD maps by first transforming sensory data into BEV space. A key challenge in this process is to extract relevant information from camera images and map it to the BEV space. %
In this work, we address this challenge by designing a probabilistic formulation that guides geometric mapping to capture map-specific elements from camera images for accurate HD mapping.

The first approaches to online HD map estimation focused on switching from rasterized to vectorized~\cite{Qi2022ICRA} and improving vectorized prediction with architectural modifications and additional losses~\cite{Yicheng2023ICML, Bencheng2023ICLR, Bencheng2024IJCV}.
Recent approaches focus on incorporating temporal information \cite{Tianyuan2024WACV, Nan2025WACV, Jiacheng2024ECCV, Jingyu2024ARXIV}, where the selection of relevant information becomes more crucial due to the increased information with the time axis. However, none of these approaches focus on improving the accuracy of the transformation and rely on standard methods, such as depth-based~\cite {Jonah2020ECCV} or attention-based~\cite{Zhiqi2022ECCV, Shaoyu2022ARXIV}, to project image features to the BEV space. We first show that replacing the attention-based projection in the SOTA method~\cite{Jiacheng2024ECCV} with a simple projection~\cite{Harley2023ICRA}, \ie, sampling the corresponding image features based on camera parameters, performs surprisingly well, even slightly outperforming the original approach.

We refer to this mapping based on camera parameters as static mapping and highlight its shortcomings for HD maps, such as inaccuracies when the slope changes or when the road is occluded by scene elements like vehicles or pedestrians. Our key idea is that adapting the mapping to the scene can enhance the transformation by more precisely extracting relevant information. However, fully relying on data is also not ideal, as evidenced by the increased false positives in attention-based mapping approaches. Our novelty lies in a probabilistic projection approach for HD mapping. %
We learn probabilistic distributions and associated confidences over sampling locations to (i) adjust mapping offsets and (ii) filter out irrelevant elements that should not influence the HD map. We also improve temporal processing by using predicted confidences to selectively accumulate reliable information over time.

We evaluated our approach on the challenging new splits of nuScenes and Argoverse2, ensuring separate training and test regions for generalization. Our results show that probabilistic projection significantly outperforms the state-of-the-art across nearly all metrics on both datasets. In particular, we observe improvements in challenging long-range perception on both datasets. While temporal consistency remains similar to the state-of-the-art on Argoverse2, it is significantly improved on nuScenes. 
Our visual analysis reveals fewer false positives, especially in pedestrian crossings, compared to attention-based baselines. %
Our contributions can be summarized as follows: 
\begin{itemize}
\item A novel probabilistic projection method to improve mapping from camera images to BEV by better capturing road structures.
\item Improved temporal processing with confidence-based selection, effectively accumulating historical information.
\item Advancing state-of-the-art in both short and long perception ranges on nuScenes and Argoverse2.
\end{itemize}

\section{Related Work}
\label{sec:rw}
\boldparagraph{Online Vectorized HD Mapping} Recent HD map methods build maps online from multiple camera views, using vector representations that preserve instance details. HDMapNet~\cite{Qi2022ICRA} is the first method to frame the vectorized HD mapping task as a rasterized BEV segmentation problem. However, it requires post-processing to convert the resulting semantic, instance, and directional BEV maps into a fully vectorized map representation.
VectorMapNet~\cite{Yicheng2023ICML} introduces an end-to-end framework that directly predicts polylines with a DETR-like detection head~\cite{Nicolas2020ECCV}, but it generates polyline vertices in an autoregressive manner, which may cause inefficiency. MapTR~\cite{Bencheng2023ICLR} has a similar but more efficient transformer-based architecture, using hierarchical modeling for map elements and matching queries level by level to predict all vertices of an element at once. MapTRv2~\cite{Bencheng2024IJCV} improves MapTR with auxiliary segmentation losses and a memory-efficient transformer decoder using decoupled self-attention. Some works focus on refining the interactions in the decoder mechanism and/or adding additional task losses~\cite{Zihao2024ECCV, Zhenhua2024ECCV, Haotian2024ECCV, Xiaolu2024CVPR}. Different from these approaches, Mask2Map~\cite{Sehwan2024ECCV} replaces the detection-based pipeline with a segmentation mask-based approach. Others concentrate on improving vectorized map modeling \cite{Wenjie2023ICCV, Limeng2023CVPR, Zhixin2024ECCV, Yi2024CVPR}.

\boldparagraph{Incorporating Temporal Information} 
StreamMapNet~\cite{Tianyuan2024WACV} uses a memory module with a streaming strategy~\cite{Chunrui2024RAL, Shihao2023ICCV} to incorporate temporal information, processing each frame separately while propagating a hidden state to maintain long-term temporal connections. SQD-MapNet~\cite{Shuo2024ECCV} inserts denoised ground truth via queries to improve the temporal consistency of map elements. PrevPredMap~\cite{Nan2025WACV} selects the top predicted queries based on their confidence to form new queries with category and location information. MapTracker~\cite{Jiacheng2024ECCV} uses two strategies for its BEV and vector memory modules. For its BEV memory module, it selects past information based on distance rather than merging all historical data, reducing information loss. %
For its vector module, it uses query propagation from tracking to link elements from previous frames with those in the current frame. MemFusionMap~\cite{Jingyu2024ARXIV} utilizes a temporal overlap heatmap to combine the current and past features, highlighting areas that should favor current data over memory while also encoding vehicle trajectory for better temporal reasoning. Lastly, MapUnveiler~\cite{Nayeon2024NEURIPS} utilizes clip-level information from adjacent frames by learning tokens that help with occluded elements, ensuring long-term consistency.

Recent work focuses on architectural changes~\cite{Yicheng2023ICML, Bencheng2023ICLR, Bencheng2024IJCV, Zihao2024ECCV, Zhenhua2024ECCV, Haotian2024ECCV, Xiaolu2024CVPR} or temporal module~\cite{Tianyuan2024WACV, Nan2025WACV, Jiacheng2024ECCV, Jingyu2024ARXIV, Nayeon2024NEURIPS, Shuo2024ECCV} while relying on off-the-shelf methods~\cite{Jonah2020ECCV, Junjie2022ARXIV, Zhiqi2022ECCV, Shaoyu2022ARXIV} for projecting image features into BEV. In this work, we focus on improving the projection with a probabilistic framework to capture map elements more accurately. For the vector module, we build on MapTracker's design~\cite{Jiacheng2024ECCV}, which is shown to achieve the best performance.

\begin{figure}[t!]
    \centering
    \includegraphics[width=0.90\linewidth]{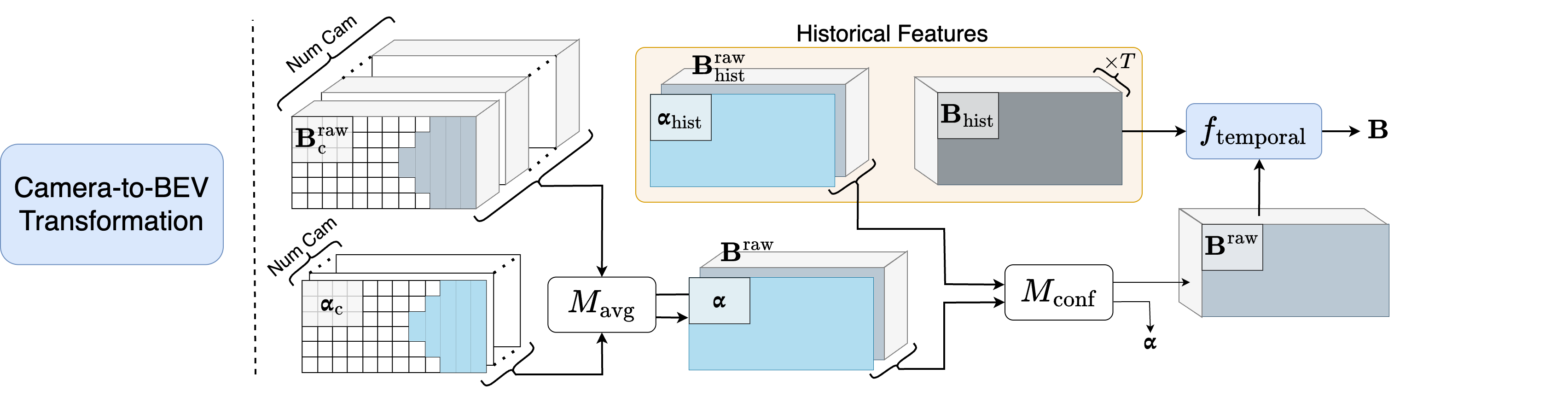}
\vspace{4pt}

\caption{\textbf{Overall Framework.}
We start by projecting camera features onto the BEV grid using probabilistic projection (detailed in \figref{fig:prob_proj}), resulting in per-camera raw BEV features and the corresponding confidence scores $\{\bB^{\text{raw}}_c, \balpha_c \mid c \in \mathcal{C}\}$ for each camera. 
Then, we merge per-camera features and confidences ($M_{\text{avg}}$), resulting in $\bB^{\text{raw}}$. 
We then incorporate temporal information by taking a weighted sum ($M_{\text{conf}}$) with historical raw BEV features $\bB^{\text{raw}}_{\text{hist}}$ according to the confidence scores. 
Finally, we process the raw BEV features and the historical BEV features $\bB_{\text{hist}}$ with $f_{\text{temporal}}$ to obtain the final BEV features, $\bB$.}
\label{fig:overall}
\end{figure}

\boldparagraph{Bird's Eye View Segmentation} 
Techniques used in BEV perception are essential for map learning since they transform 2D camera features into the BEV space. This transformation can be divided into three approaches. \textit{Depth-based}~\cite{Jonah2020ECCV, Junjie2022ARXIV} models learn an explicit depth distribution at each pixel. \textit{Attention-based} methods~\cite{Brady2022CVPR, Shaoyu2022ARXIV} %
learn geometry implicitly through cross-attention between sensor features and predefined 3D grid coordinates. \textit{Sampling-based}~\cite{Harley2023ICRA, Loick2024CVPR} methods map voxel grids to camera image pixels and bilinearly sample the intersecting image features. %

\begin{figure*}[t]
    \centering
    \includegraphics[width=0.9\linewidth]{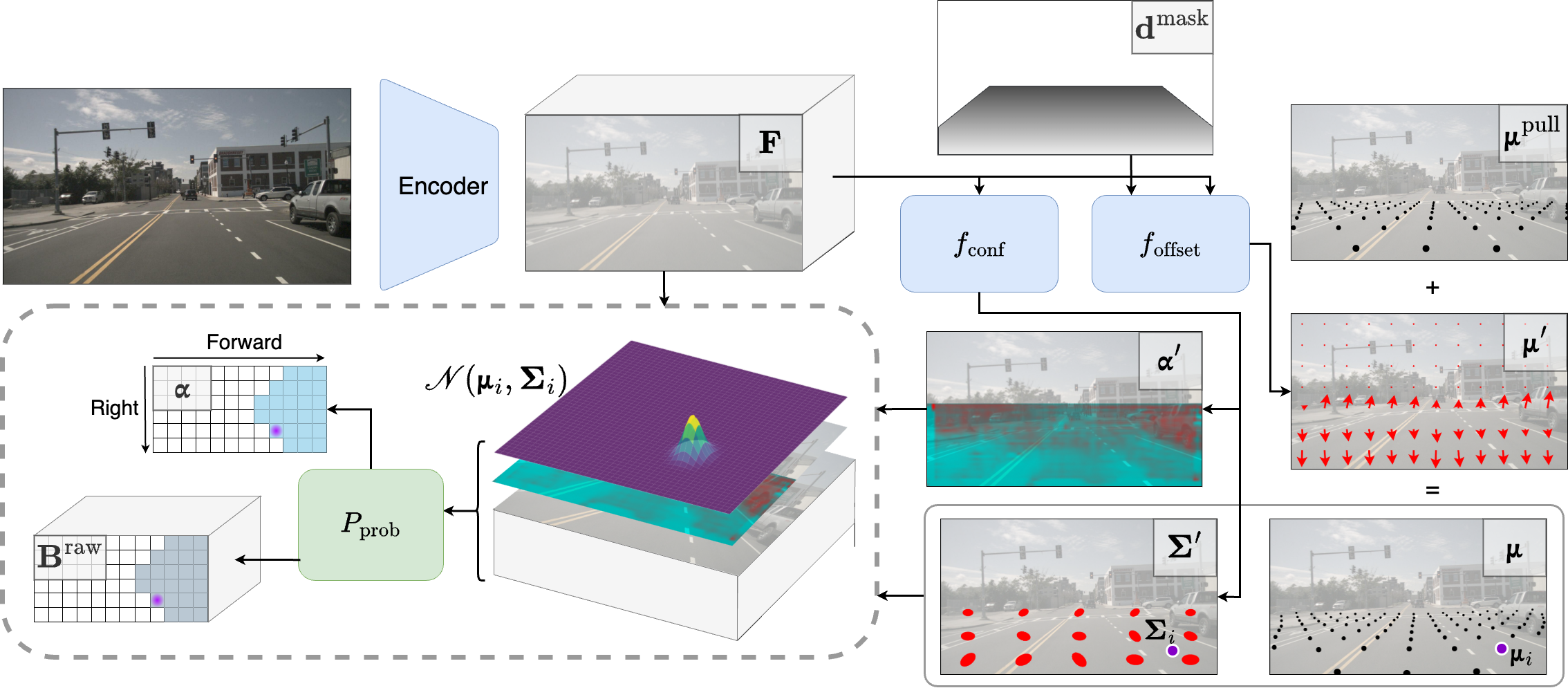}
    \vspace{7pt} 
    \caption{\textbf{Camera-to-BEV Transformation using Probabilistic Projection.} We illustrate the camera-to-BEV transformation for a single camera by dropping the subscript $c$ in the notation. The camera image is first processed by an encoder to extract features $\bF$. These features are then processed by two networks, $f_\text{offset}$ and $f_\text{conf}$, to predict the probabilistic projection parameters. Given a distance mask $\bd^\text{mask}$ as an additional input, $f_\text{offset}$ predicts the offset values $\bmu^\prime$, which adjust the static BEV-to-image mapping $\bmu^\text{pull}$ to form a more accurate mapping $\bmu$. Meanwhile, $f_\text{conf}$ estimates the covariance $\bSigma^\prime$ and the confidence scores $\balpha^\prime$. 
    \textbf{Probabilistic Projection (dashed box):} For a grid cell $i$, the corresponding values $\bmu_i$ and $\bSigma_i$ from $\bmu$ and $\bSigma^\prime$ parameterize the Gaussian distribution $\cN(\bmu_i, \bSigma_i)$. Based on this distribution, we sample $K$ locations in the image and construct the feature of the grid cell as a weighted sum of feature vectors projected from these sampled locations. The weight, representing each pixel's contribution, is determined by the likelihood of the mapping and the pixel's confidence. Note that we use $\prime$ with a variable to denote the variable in image space, except for $\bmu^\text{pull}$ and $\bmu$, which are shown in image space for illustration purposes but are formally defined in BEV space. For more details, see the text.}
    \label{fig:prob_proj}
\end{figure*}

Both depth-based~\cite{Jonah2020ECCV} and attention-based~\cite{Shaoyu2022ARXIV} models have been utilized by the vectorized HD mapping methods~\cite{Bencheng2023ICLR, Bencheng2024IJCV}. However, depth-based methods struggle with the complexity of accurately predicting depth for every pixel, while attention-based methods do not benefit from explicit geometric modeling. Both approaches are prone to projection errors. BEVFormer~\cite{Zhiqi2022ECCV} uses both attention and sampling mechanisms with deformable attention to aggregate image features into 3D reference grids, a method preferred by the latest mapping methods~\cite{Tianyuan2024WACV, Jiacheng2024ECCV}. However, this method suffers from false positives by placing road elements where they occur most frequently in the training data distribution. With the proposed probabilistic projection for HD mapping, our method better preserves the geometry of road structures.

\section{Methodology}
\label{sec:method}
Given images from multiple cameras surrounding a vehicle, online mapping approaches aim to extract an HD map of the environment by segmenting lane boundaries, dividers, and pedestrian crossings in BEV representation. Recent work~\cite{Jiacheng2024ECCV} first constructs a BEV feature space using a BEV encoder. Based on the extracted BEV features, a vectorized map representation is decoded. In this work, we focus on improving the BEV feature space in the first stage by introducing a novel probabilistic projection mechanism. 

Our overall framework is illustrated in \figref{fig:overall}, with the corresponding steps detailed in \algref{algo}. Additionally, the probabilistic projection is illustrated in \figref{fig:prob_proj}.

We first process camera images with a ResNet~\cite{Kaiming2016CVPR} encoder, followed by a Feature Pyramid Network, resulting in features $\bF_c \in \nR^{C \times H \times W}$ for the camera $c$ where $H \times W$ denotes the reduced spatial resolution of the image and $C$, the hidden dimension size. 
Assuming known intrinsics and extrinsics, we can define the mapping $\bmu^{\text{pull}}_c \in \nR^{h \times w \times 2}$ to pull features from the image plane of the camera $c$ to a BEV grid of size $h \times w $.

\subsection{Probabilistic Projection} 
The static mapping $\bmu^{\text{pull}}_c$ based on camera parameters fails to capture scene structures, \eg, when the slope changes (see Supplementary for a visualization). In this work, we propose to improve this mapping by predicting per-pixel offset distributions $\cN(\bmu^{\prime}_c, \bSigma_c^{\prime})$ and an associated confidence map $\balpha_c^{\prime}$ to direct the offsets towards more relevant locations on the image:
\begin{eqnarray}
        \bmu^{\prime}_c &=& f_{\text{offset}}(\bF_c, \bd^{\text{mask}}_c) \\
        \bSigma_c^{\prime}, \balpha_c^{\prime} &=& f_{\text{conf}}(\bF_c) \label{eq:sigma_alpha_pred}
\end{eqnarray}
where $f_{\text{offset}}$ is a 3-layer CNN processing the camera features $\bF_c$ together with a distance mask $\bd^{\text{mask}}_c \in \mathbb{R}^{H \times W}$ for positional information. Since the correction we want to apply to a point depends on how far the point is from the camera due to perspective projection, we encode the distance mask $\bd^{\text{mask}}_c$ with a 3-layer CNN and provide it as input to the offset predictor $f_{\text{offset}}$.
We use another 3-layer CNN, $f_{\text{conf}}$, to predict covariance and confidences; differently, we do not provide the distance mask as input. %

\begin{algorithm}[t]
\caption{Map Prediction with Probabilistic Projection}
\label{algo}
\begin{algorithmic}[1]
\footnotesize
\STATE \textbf{Input:} The set of cameras $\mathcal{C}$, combined historical raw BEV features $\bB^{\text{raw}}_\text{hist}$, their confidence scores $\balpha_{\text{hist}}$, selected historical BEV features $\bB_{\text{hist}}$, and for each camera, encoded features, distance mask, and pull indices, $\{\bF_c, \bd^{\text{mask}}_c, \bmu^{\text{pull}}_c \vert c \in \cC \}$

\STATE \textbf{Output:} Predicted BEV features, $\bB$

\begin{multicols}{2} %

\FOR{each camera $c \in \cC$}

    \STATE \textbf{Offset Prediction}\\
    $\bmu^{\prime}_c \gets f_{\text{offset}}(\bF_c, \bd^{\text{mask}}_c)$

    \STATE \textbf{Covariance and Confidence Prediction}\\
    $(\bSigma_c^{\prime}, \balpha_c^{\prime}) \gets f_{\text{conf}}(\bF_c)$

    \STATE \textbf{Offset Sampling}\\
    $\bmu^{\text{off}}_c \gets P(\bmu^{\prime}_c, \bmu^{\text{pull}}_c)$

    \STATE \textbf{Updating Pull Index}\\
    $\bmu_c \gets \bmu^{\text{pull}}_c + \bmu^{\text{off}}_c$

    \STATE \textbf{Covariance Sampling}\\
    $\bSigma_c \gets P(\bSigma_c^{\prime}, \bmu_c)$

    \STATE \textbf{Probabilistic Projection}\\
    $(\bB^{\text{raw}}_c, \balpha_c) \gets P_{\text{prob}}(\bF_c, \bmu_c, \bSigma_c, \balpha_c^{\prime})$

\ENDFOR

\columnbreak 

\STATE \textbf{Merging Camera Features} \\ $\bB^{\text{raw}}, \balpha \gets M_\text{avg}(\left\{ \bB^{\text{raw}}_c \mid c \in \mathcal{C} \right\}), M_\text{avg}(\left\{ \balpha_c \mid c \in \mathcal{C} \right\})$
\\
\STATE \textbf{Merging with Historical Raw Features}\\
$(\bB^{\text{raw}}, \balpha) \gets M_\text{conf}(\bB^{\text{raw}}, \balpha, \bB^{\text{raw}}_\text{hist}, \balpha_{\text{hist}})$

\STATE \textbf{Parametric Temporal Fusion}\\
$\bB \gets f_{\text{temporal}}(\bB^{\text{raw}}, \bB_{\text{hist}}) $

\STATE \textbf{Save} $\bB$, $\bB^{\text{raw}}$, $\balpha$

\STATE \textbf{Return} $\bB$
\end{multicols} 
\end{algorithmic}
\end{algorithm}

\boldparagraph{Updated Projection Parameters}
Given the projection parameters, \ie offsets $\bmu^{\prime}_c$ and the covariance $\bSigma_c^{\prime}$, predicted from the camera features, we project them to the BEV grid. The projection operation, denoted by $P(\cdot, \cdot)$, uses the second argument to sample bilinearly from the first argument:

\begin{eqnarray}
    \label{eq:mu_proj}
    \bmu^{\text{off}}_c &=& P(\bmu^{\prime}_c, \bmu^{\text{pull}}_c) \\
    \label{eq:off_res}
    \bmu_c &=& \bmu^{\text{pull}}_c + \bmu^{\text{off}}_c \\
    \label{eq:sigma_proj}
    \bSigma_c &=& P(\bSigma_c^{\prime}, \bmu_c)
\end{eqnarray}

We first project offsets $\bmu^{\prime}_c$ using the static mapping $\bmu^{\text{pull}}_c$ in \eqref{eq:mu_proj}.
We then residually add the projected offsets $\bmu^{\text{off}}_c$ to the static mapping $\bmu^{\text{pull}}_c$  to obtain the updated mappings $\bmu_c$ in \eqref{eq:off_res}. Finally, we project the covariance $\bSigma_c^{\prime}$ by using the updated mappings $\bmu_c$ in \eqref{eq:sigma_proj}.

On a practical note, the projection is implemented using \texttt{grid\_sample}, followed by a \texttt{permute} operation to arrange dimensions. To further clarify the notation and dimensionality, we use $\bmu_c^{\prime}, \bSigma_c^{\prime}$ to represent the probabilistic projection parameters with respect to the perspective camera at the spatial resolution $H \times W$. We denote the projected version respectively as $\bmu_c, \bSigma_c$, in the spatial resolution of the BEV grid, \ie $h \times w$. 
The offset parameters $\bmu_c$ have 2 channels and the covariance $\bSigma_c$ has 3 channels. %
Note that $\bmu_c$ and $\bSigma_c$ are empty in the locations where the corresponding BEV grid is outside the camera's field of view.

\boldparagraph{Probabilistic Feature Projection ($P_{\text{prob}}$)}
The updated projection parameters define a probabilistic mapping at each grid cell, pointing to the most likely locations in the respective camera to pull features from. 
Based on this distribution, we sample $K$ pixels from camera features $\bF_c$ for a grid cell and then represent the grid cell as a weighted sum of feature vectors of the sampled pixels. We define the weight by combining the Gaussian likelihoods and the confidence scores of the sampled locations.

Formally, let $\bB^{\text{raw}}_c \in \nR^{C \times h \times w}$ be the result of the probabilistic projection of the image features for the camera $c$. For a grid cell $i$ on $\bB^{\text{raw}}_c$, let $\bb_i \in \nR^{C\times 1}$ denote its feature vector pulled from camera $c$ according to the probabilistic mapping proposed. Dropping the dependency on camera index $c$ for clarity, we first sample $K$ locations based on the distribution $\cN(\bmu_i, \bSigma_i)$ of the grid cell $i$. We then construct $\bb_i$ as a weighted sum of feature vectors projected from the sampled locations. The weight $w_k$ denotes the contribution of pixel $k$. We define it based on the predicted confidence $\balpha_k^\prime$ for the pixel $k$ \eqref{eq:sigma_alpha_pred} and the likelihood of the mapping.

\begin{equation}\label{eq:pp_wsumANDpp_w}
\begin{gathered}
    \begin{aligned}
        \bmu_{i,k} &\sim \cN(\bmu_i, \bSigma_i) \\
        \bb_i &= \sum_{k=1}^K w_k~P(\bF, \bmu_{i,k})
    \end{aligned}
    \hspace{2cm}
    w_k = \balpha_k^\prime \cdot \cN(\bmu_{i,k}; \bmu_i, \bSigma_i)
\end{gathered}
\end{equation}

We normalize the likelihood of each sample $k$, $\cN(\bmu_{i,k}; \bmu_i, \bSigma_i)$, by the sum of $K$ Gaussian likelihoods. 
The confidence score for each grid location $i$ is updated as the sum of weights over sampled locations $\alpha_i = \sum_{k=1}^K w_k$.

\boldparagraph{Merging Camera Features} We merge the resulting BEV grid  $\bB^\text{raw}_c$ and $\balpha_c$ from each camera into $\bB^{\text{raw}}$ and $\balpha$ by a simple averaging ($M_{\text{avg}}$): 
\begin{equation}
\begin{aligned}
    \bB^\text{raw} &= \sum_{c} \bB^\text{raw}_c \oslash \max(1, \bV)
\end{aligned}
\hspace{2cm}
\begin{aligned}
    \balpha &= \sum_{c} \balpha_c \oslash \max(1, \bV)
\end{aligned}
\end{equation}
where $\oslash$ denotes the elementwise Hadamard division and $\bV \in \{0, \dots, C\}^{h \times w}$ counts the number of cameras contributing to each grid cell.

\subsection{Temporal Information}
Previous works \cite{Nan2025WACV, Jiacheng2024ECCV} have shown the importance of utilizing temporal information for HD map prediction. For example, MapTracker~\cite{Jiacheng2024ECCV} warps the previous BEV estimation to use as initialization at the current time step and proposes a memory mechanism. We also utilize temporal information in two ways.

\boldparagraph{Historical Raw Features} First, we keep track of the most confident raw features and their associated confidences for each grid cell over a time horizon in $\bB^\text{raw}_\text{hist}$ and $\balpha_\text{hist}$. We update $\bB^\text{raw}$ according to historical information by first warping $\bB^\text{raw}_\text{hist}$ to align with the current time step. The update is performed simply by  adding the two, each weighted by their corresponding confidences ($M_{\text{conf}}$). We update the confidence matrix $\balpha$ in a similar way (see Supplementary for details).

\begin{equation}\label{eq:hist_rawANDdef}
\begin{aligned}
\bB^\text{raw} &= \balpha \odot \bB^\text{raw} + \balpha_\text{hist} \odot \bB^\text{raw}_\text{hist} \\
\bB^\text{raw} &= \bB^\text{raw} \oslash (\balpha + \balpha_\text{hist})
\end{aligned}
\hspace{2cm}
\begin{aligned}
\balpha &= \balpha \odot \balpha + \balpha_\text{hist} \odot \balpha_\text{hist} \\
\balpha &= \balpha \oslash (\balpha + \balpha_\text{hist})
\end{aligned}
\end{equation}

We normalize both by dividing elementwise by the sum of confidences. We save them in the history buffer to be warped and used in future time steps.

\boldparagraph{Parametric Temporal Fusion ($f_{\text{temporal}}$)}
So far, the merged BEV grid is simply the result of projecting image features to the BEV grid, \ie \emph{raw}, without any learning on top of it. Following the memory mechanism proposed in MapTracker~\cite{Jiacheng2024ECCV}, we store the last 20 estimated BEV representations in memory and select a subset of size $T=4$ based on vehicle positions, denoted by $\bB_\text{hist} \in \nR^{T \times C \times h \times w}$. We concatenate $\bB^\text{raw}$ and $\bB_\text{hist}$ and feed it into $f_{\text{temporal}}$ to obtain the final BEV feature map, $\bB$.
\begin{table*}[th!]
  \centering
  \footnotesize %
  \setlength{\tabcolsep}{2.5pt}
  \caption{\textbf{Quantitative Results.} Comparison with the state-of-the-art models on nuScenes~\cite{Holger2020CVPR} and Argoverse2~\cite{Benjamin2023ARXIV} in two ranges, using the new split. Our method outperforms existing methods on both datasets in almost all metrics, with significant improvements on nuScenes. %
  }
  \vspace{10pt}
\begin{tabular}{llllllll}
    \toprule
      \textbf{Range~(m)} & 
      \textbf{Dataset} & 
      \textbf{Method} & 
      ${AP_p}$ & 
      ${AP_d}$ & 
      ${AP_b}$ & 
      $mAP$ & 
      $C$-$mAP$ \\
    \midrule       
    \multirow{6}{*}{60$\times$30}
        & \multirow{3}{*}{nuScenes ~\cite{Holger2020CVPR}} & 
            StreamMapNet~\cite{Tianyuan2024WACV} & 31.6 & 28.1 & 40.7 & 33.5 & 22.2  \\
            & & MapTracker~\cite{Jiacheng2024ECCV}   & 45.9 & 30.0 & 45.1 & 40.3 & 32.5 \\
            & & Ours         & 49.8 \color{blue}{\tiny(+8.5\%)} 
                             & 36.2  \color{blue}{\tiny(+20.7\%)} 
                             & 50.1  \color{blue}{\tiny(+11.1\%)} 
                             & 45.4  \color{blue}{\tiny(+12.7\%)} 
                             & 37.2  \color{blue}{\tiny(+14.5\%)} \\
        \cmidrule(lr){2-8} 
        & \multirow{3}{*}{Argoverse2 ~\cite{Benjamin2023ARXIV}} & 
            StreamMapNet~\cite{Tianyuan2024WACV} & 61.8 & 68.2 & 63.2 & 64.4 & 54.4  \\
            & & MapTracker~\cite{Jiacheng2024ECCV}   & 70.0 & 75.1 & 68.9 & 71.3 & 63.2 \\
            & & Ours         & 72.9
                            \color{blue}{\tiny(+4.1\%)}
                             & 76.9 
                             \color{blue}{\tiny(+2.4\%)}
                             & 67.5 
                             \color{red}{\tiny(-2.0\%)}
                             & 72.4               \color{blue}{\tiny(+1.5\%)}
                             & 62.7  \color{red}{\tiny(-0.8\%)} \\
    \midrule
    \multirow{6}{*}{100$\times$50} 
        & \multirow{3}{*}{nuScenes ~\cite{Holger2020CVPR}} & 
            StreamMapNet~\cite{Tianyuan2024WACV} & 25.1 & 18.9 & 25.0 & 23.0 & 14.6  \\ 
            & & MapTracker~\cite{Jiacheng2024ECCV}  & 45.9 & 24.3 & 38.4 & 36.2 & 27.5 \\
            & & Ours         & 52.5 \color{blue}{\tiny(+14.4\%)} 
                             & 33.1  \color{blue}{\tiny(+36.2\%)} 
                             & 42.9  \color{blue}{\tiny(+11.7\%)} 
                             & 42.8  \color{blue}{\tiny(+18.2\%)} 
                             & 33.5  \color{blue}{\tiny(+21.8\%)} \\
        \cmidrule(lr){2-8} 
        & \multirow{3}{*}{Argoverse2 ~\cite{Benjamin2023ARXIV}} & 
            StreamMapNet~\cite{Tianyuan2024WACV} & 60.1 & 56.1 & 47.5 & 54.6 & 41.3  \\
            & & MapTracker~\cite{Jiacheng2024ECCV}   & 71.2 & 64.6 & 58.5 & 64.8 & 55.7 \\
            & & Ours         & 74.9                     
                             \color{blue}{\tiny(+5.2\%)}
                             & 67.4 
                              \color{blue}{\tiny(+4.3\%)}
                             & 58.9 
                              \color{blue}{\tiny(+0.7\%)}
                             & 67.1          \color{blue}{\tiny(+3.5\%)}
                             & 56.1
                              \color{blue}{\tiny(+0.7\%)} \\
    \bottomrule
  \end{tabular}
  \label{tab:main}
\end{table*}
\section{Experiments}
\label{sec:exp}
\boldparagraph{Experimental Setup} 
We evaluate our model on the nuScenes~\cite{Holger2020CVPR} and Argoverse2~\cite{Benjamin2023ARXIV} datasets, using the revised non-overlapping geographical splits proposed in StreamMapNet~\cite{Tianyuan2024WACV} to better assess generalization, and adopt MapTracker’s~\cite{Jiacheng2024ECCV} temporally-aligned ground truth. Performance is measured under short (60m$\times$30m) and long (100m$\times$50m) range settings using Average Precision (AP) at varying distance thresholds, and reported per class as $AP_p$, $AP_d$, and $AP_b$, with both mean AP ($mAP$) and consistency-aware $C$-$mAP$. We follow the MapTracker~\cite{Jiacheng2024ECCV} training pipeline for both datasets. Full details are provided in the Supplementary.

\subsection{Quantitative Comparison}
We compare our approach with state-of-the-art online mapping methods StreamMapNet~\cite{Tianyuan2024WACV} and MapTracker~\cite{Jiacheng2024ECCV} in \tabref{tab:main}. We directly obtained the baseline results from MapTracker and verified its performance using the official codebase, while the StreamMapNet results are taken as reported. Improvements over the previous best results are highlighted in green as percentages, while performance losses are shown in red.

We first note the significant improvement in generalization capability with our method, as the new split ensures no overlapping regions between the training and test sets. The improvements are consistent across almost all categories on both datasets, except for the boundary class on Argoverse2. The results are particularly impressive on nuScenes, with a 12.7\% overall improvement in mAP in the short range and 18.2\% in the long range, consistently across all categories, especially the divider. %
The largest improvements on both datasets are achieved in the long range, with +18.2\% on nuScenes and +3.5\% on Argoverse2. This is particularly noteworthy given the difficulty of long perception ranges, such as distant pedestrian crossings or long dividers. 
Furthermore, our method improves the temporal consistency on nuScenes in both ranges, demonstrating the effectiveness of temporal processing in our approach. 

\boldparagraph{nuScenes \vs Argoverse2} Our improvements are more pronounced on nuScenes than on Argoverse2. In general, all methods perform better on Argoverse2 than on nuScenes, probably due to its better diversity, covering six cities compared to two on nuScenes. Furthermore, the longer viewing range of the cameras in Argoverse2 reduces the gap between the short- and long-range perception results compared to nuScenes. On Argoverse2, our method achieves notable improvements in pedestrian crossings and lane dividers across both ranges. However, while boundary detection improves slightly in the long range, it performs worse than MapTracker in the short range. Similarly, $C$-$mAP$ decreases slightly in the short range but shows a slight improvement in the long range. 

\subsection{Qualitative Comparison}
\begin{figure*}[ht]
    \centering
    \bmvaHangBox{\parbox{1.0\textwidth}{~\\[2mm] 
    \hspace{0mm}\includegraphics[width=\textwidth]{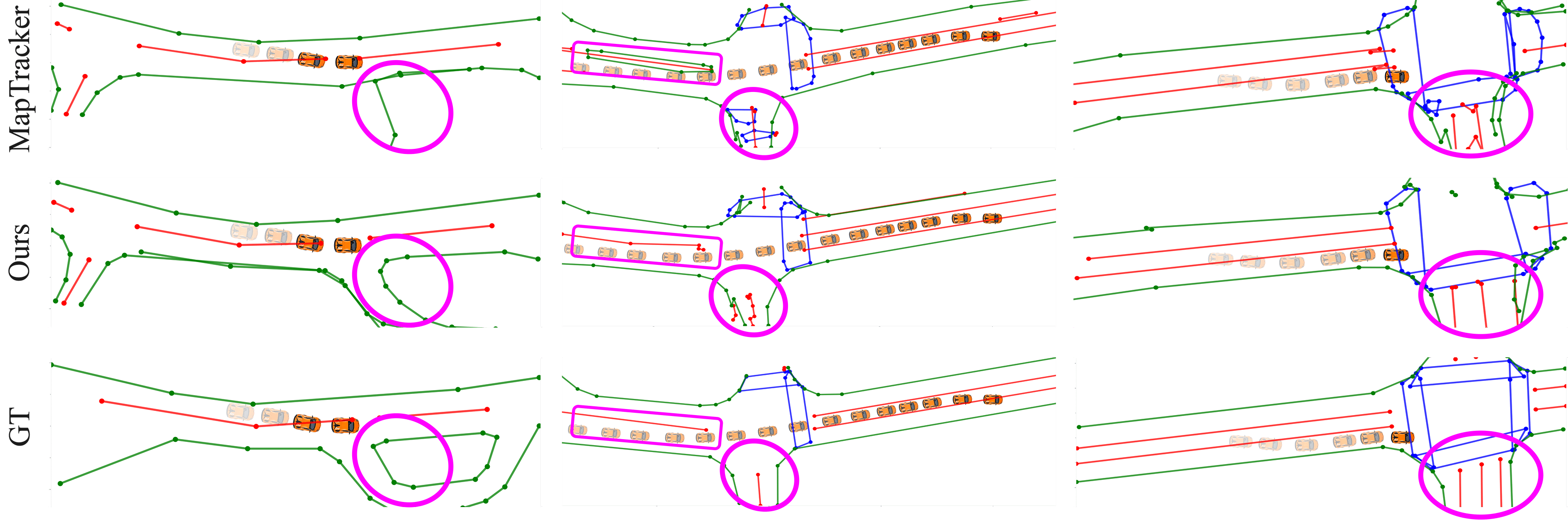}\\[0.1pt]}}  %
    \vspace{3mm}  %
    \caption{\textbf{Qualitative Comparison.} We qualitatively compare our method to MapTracker~\cite{Jiacheng2024ECCV}. Pedestrian crossings, lane dividers, and road boundaries are shown in blue, red, and green, respectively. Our improvements are highlighted with a purple circle for road boundaries (left), pedestrian crossings (middle), and lane dividers (right) on nuScenes.}
    \label{fig:qual_final}

\end{figure*}

In \figref{fig:qual_final}, we present qualitative comparison of our method with MapTracker~\cite{Jiacheng2024ECCV}. While MapTracker often introduces false positives, particularly by placing pedestrian crossings near road junctions (middle), where they frequently appear in the training data, our geometry-based mapping avoids such mode-fitting behavior. Additionally, as shown in \figref{fig:qual_final} (right and left), our approach more faithfully captures the actual structure of road elements, producing more accurate representations of lane dividers and boundaries. Please see Supplementary for more qualitative analysis.

\subsection{Ablation Study}
In \tabref{tab:component_ablation}, we evaluate the effect of the proposed components on nuScenes in the short range. We begin with the vanilla version in $A$, which replaces the deformable attention-based projection~\cite{Zhiqi2022ECCV} in MapTracker~\cite{Jiacheng2024ECCV} with a static pull mechanism using bilinear sampling~\cite{Harley2023ICRA}. Note that the vanilla version still has the vector module and the memory mechanism from MapTracker. This version already slightly outperforms MapTracker (40.3 \vs 41.0), showing the potential of a geometry-based projection~\cite{Harley2023ICRA}.

\begin{table}[b]
\centering
\footnotesize
\setlength{\tabcolsep}{3pt}
\caption{\textbf{Ablation Study.} We evaluate the effect of the proposed components: probabilistic projection ($\cN(\bmu, \bSigma)$), confidence scores ($\balpha$), and historical raw features ($\bB_{\text{hist}}^{\text{raw}}$) on nuScenes in the short range.}
\vspace{4pt}
\begin{tabular}{c ccc ccccc}
    \toprule
    \cmidrule(lr){1-3} 
    \cmidrule(lr){4-8}
    ID &
    $\cN(\bmu, \bSigma)$ &
    $\balpha$ & 
    $\bB_{\text{hist}}^{\text{raw}}$ & 
    ${AP_p}$ & 
    ${AP_d}$ & 
    ${AP_b}$ & 
    $mAP$ & 
    $C$-$mAP$ \\
    \midrule
    $A$ & \cross & \cross & \cross & 43.9 & 31.6 & 47.4 & 41.0 & 33.4 \\
    \cmidrule{1-9}
    $B_1$ & \multirow{2}{*}{\tick} & \cross & \multirow{2}{*}{\cross}    & 48.8 & 35.2 & 46.6 & 43.5 & 34.9 \\
    $B_2$ &  & \tick &  & 48.9 & 34.2 & 48.0 & 43.6 & 35.4 \\
    \cmidrule{1-9}
    $C_1$ & \multirow{2}{*}{\cross}  & \cross & \multirow{2}{*}{\tick}  & 44.2 & 32.6 & 46.0 & 41.0 & 33.2 \\
    $C_2$ &  & \tick &   & 46.6 & 35.6 & 48.1 & 43.4 & 35.2 \\
    \cmidrule{1-9}
    $D$ & \multirow{2}{*}{\tick} & \cross & \multirow{2}{*}{\tick}   & 47.4 & 34.2 & 48.6 & 43.4 & 35.3 \\
    $E$ & & \tick &    & 49.8 & 36.2 & 50.1 & 45.4 & 37.2 \\
    \bottomrule
\end{tabular}
\label{tab:component_ablation}
\end{table}

\boldparagraph{Probabilistic Projection ($B_{1,2}$)} Learning to adjust the static mapping with per-pixel offset distributions improves the performance from 41.0 to 43.5 ($B_1$). These results support our motivation to update projection parameters based on scene structures. Grid points adapt to road slope changes, improving coverage of pedestrian crossings and road boundaries (see Supplementary for a visualization).

Results do not change significantly with the inclusion of confidences ($B_2$), likely due to the uncertainty already being captured with $\bSigma$. Without $\bB_{\text{hist}}^{\text{raw}}$ in experiments $B_{1,2}$, the confidences do not play any other role in the pipeline. %

\boldparagraph{Historical Raw Features ($C_{1,2}$)} Merging historical raw features without using confidence scores ($C_1$) does not lead to any improvement. However, when historical information is selected based on confidence scores ($C_2$), $mAP$ increases by +2.4, highlighting the importance of selective merging. Additionally, in $C_2$, $C$-$mAP$ also improves, indicating better temporal consistency with raw historical features, weighted by confidence. Note that these additional gains come from temporal processing on top of the memory mechanism of MapTracker, which selects frames based on the positions of the vehicle. %
With confidence scores, our raw historical features selectively accumulate temporal information, keeping only the most reliable information, which leads to improved performance. %

\boldparagraph{Confidence Map ($D, E$)} We use confidence scores in two ways: for weighting features in weighted projection \eqref{eq:pp_wsumANDpp_w}, and for selecting and merging historical raw BEV features into the current \eqref{eq:hist_rawANDdef}. 
Removing confidence scores has the least impact on $mAP$ ($D$), likely because uncertainty in weighted projection can still be captured by $\bSigma$ as mentioned earlier. 
However, confidence-based selection still plays a crucial role in merging historical raw features, as the best performance is achieved with the complete model ($E$) when confidences are included. %
The confidence score learns to assign high values to road structures while filtering out pedestrians, sidewalks, walls, and buildings (see Supplementary for a visualization).

\boldparagraph{Efficiency Comparison} 
Our model, with 63.8M parameters, is similar in size to MapTracker (65.8M) and achieves a comparable inference speed of 15.9 FPS \vs 16.3 FPS for MapTracker, both measured on a single NVIDIA A100. %

\section{Conclusion}
\label{sec:conc}
We propose a probabilistic projection mechanism to enhance the accuracy of mapping from image to BEV space in online HD map estimation. Our probabilistic formulation, combined with confidence scores, significantly improves performance on the new splits of nuScenes and Argoverse2, particularly in the long range. Qualitative analysis shows that our probabilistic mapping effectively reduces false positives while adapting to scene variations. 
Additionally, our method improves temporal consistency on nuScenes through confidence-based accumulation of temporal information, emphasizing the importance of adapting temporal selection to the scene.

\boldparagraph{Limitations and Future Work}
While replacing the transformer block with an offset-based feature sampling module reduced false positives and hallucinations, we observed that the model does not fully utilize neighboring context when target regions are occluded (e.g., partially visible lane lines). Hybrid or context-aware mechanisms could be explored to address this limitation. Additionally, our findings indicate that there is room for improvement in leveraging temporal information more effectively. We leave these challenges as promising directions for future work.

\boldparagraph{Acknowledgements}
This project is funded by the European Union (ERC, ENSURE, 101116486) with additional compute support from Leonardo Booster (EuroHPC Joint Undertaking, EHPC-AI-2024A01-060). Views and opinions expressed are however those of the author(s) only and do not necessarily reflect those of the European Union or the European Research Council. Neither the European Union nor the granting authority can be held responsible for them.

\bibliography{bmvc_final_main}

\begin{thebibliography}{32}
\providecommand{\natexlab}[1]{#1}
\providecommand{\url}[1]{\texttt{#1}}
\expandafter\ifx\csname urlstyle\endcsname\relax
  \providecommand{\doi}[1]{doi: #1}\else
  \providecommand{\doi}{doi: \begingroup \urlstyle{rm}\Url}\fi

\bibitem[Caesar et~al.(2020)Caesar, Bankiti, Lang, Vora, Liong, Xu, Krishnan, Pan, Baldan, and Beijbom]{Holger2020CVPR}
Holger Caesar, Varun Bankiti, Alex~H Lang, Sourabh Vora, Venice~Erin Liong, Qiang Xu, Anush Krishnan, Yu~Pan, Giancarlo Baldan, and Oscar Beijbom.
\newblock nuscenes: A multimodal dataset for autonomous driving.
\newblock In \emph{CVPR}, 2020.

\bibitem[Carion et~al.(2020)Carion, Massa, Synnaeve, Usunier, Kirillov, and Zagoruyko]{Nicolas2020ECCV}
Nicolas Carion, Francisco Massa, Gabriel Synnaeve, Nicolas Usunier, Alexander Kirillov, and Sergey Zagoruyko.
\newblock End-to-end object detection with transformers.
\newblock In \emph{ECCV}, 2020.

\bibitem[Chambon et~al.(2024)Chambon, Zablocki, Chen, Bartoccioni, P{\'e}rez, and Cord]{Loick2024CVPR}
Loick Chambon, Eloi Zablocki, Micka{\"e}l Chen, Florent Bartoccioni, Patrick P{\'e}rez, and Matthieu Cord.
\newblock {PointBeV}: A sparse approach for bev predictions.
\newblock In \emph{CVPR}, 2024.

\bibitem[Chen et~al.(2024)Chen, Wu, Tan, Ma, and Furukawa]{Jiacheng2024ECCV}
Jiacheng Chen, Yuefan Wu, Jiaqi Tan, Hang Ma, and Yasutaka Furukawa.
\newblock Maptracker: Tracking with strided memory fusion for consistent vector hd mapping.
\newblock In \emph{ECCV}, 2024.

\bibitem[Chen et~al.(2022)Chen, Cheng, Wang, Meng, Zhang, and Liu]{Shaoyu2022ARXIV}
Shaoyu Chen, Tianheng Cheng, Xinggang Wang, Wenming Meng, Qian Zhang, and Wenyu Liu.
\newblock Efficient and robust {2D-to-BEV} representation learning via geometry-guided kernel transformer.
\newblock \emph{arXiv preprint arXiv:2206.04584}, 2022.

\bibitem[Choi et~al.(2024)Choi, Kim, Shin, and Choi]{Sehwan2024ECCV}
Sehwan Choi, Jungho Kim, Hongjae Shin, and Jun~Won Choi.
\newblock {Mask2map}: Vectorized hd map construction using bird’s eye view segmentation masks.
\newblock In \emph{ECCV}, 2024.

\bibitem[Ding et~al.(2023)Ding, Qiao, Qiu, and Zhang]{Wenjie2023ICCV}
Wenjie Ding, Limeng Qiao, Xi~Qiu, and Chi Zhang.
\newblock Pivotnet: Vectorized pivot learning for end-to-end hd map construction.
\newblock In \emph{ICCV}, 2023.

\bibitem[Han et~al.(2024)Han, Yang, Sun, Ge, Dong, Zhou, Mao, Peng, and Zhang]{Chunrui2024RAL}
Chunrui Han, Jinrong Yang, Jianjian Sun, Zheng Ge, Runpei Dong, Hongyu Zhou, Weixin Mao, Yuang Peng, and Xiangyu Zhang.
\newblock Exploring recurrent long-term temporal fusion for multi-view {3D} perception.
\newblock \emph{RAL}, 9\penalty0 (7):\penalty0 6544--6551, 2024.

\bibitem[Harley et~al.(2023)Harley, Fang, Li, Ambrus, and Fragkiadaki]{Harley2023ICRA}
Adam~W Harley, Zhaoyuan Fang, Jie Li, Rares Ambrus, and Katerina Fragkiadaki.
\newblock Simple-bev: What really matters for multi-sensor bev perception?
\newblock In \emph{ICRA}, 2023.

\bibitem[He et~al.(2016)He, Zhang, Ren, and Sun]{Kaiming2016CVPR}
Kaiming He, Xiangyu Zhang, Shaoqing Ren, and Jian Sun.
\newblock Deep residual learning for image recognition.
\newblock In \emph{CVPR}, 2016.

\bibitem[Hu et~al.(2024)Hu, Wang, Wang, Hu, Xu, and Zhang]{Haotian2024ECCV}
Haotian Hu, Fanyi Wang, Yaonong Wang, Laifeng Hu, Jingwei Xu, and Zhiwang Zhang.
\newblock {ADMap}: Anti-disturbance framework for vectorized hd map construction.
\newblock In \emph{ECCV}, 2024.

\bibitem[Huang and Huang(2022)]{Junjie2022ARXIV}
Junjie Huang and Guan Huang.
\newblock Bevpoolv2: A cutting-edge implementation of bevdet toward deployment.
\newblock \emph{arXiv preprint arXiv:2211.17111}, 2022.

\bibitem[Kim et~al.(2025)Kim, Seong, Ji, and Jang]{Nayeon2024NEURIPS}
Nayeon Kim, Hongje Seong, Daehyun Ji, and Sujin Jang.
\newblock Unveiling the hidden: Online vectorized {HD} map construction with clip-level token interaction and propagation.
\newblock In \emph{NeurIPS}, 2025.

\bibitem[Li et~al.(2022{\natexlab{a}})Li, Wang, Wang, and Zhao]{Qi2022ICRA}
Qi~Li, Yue Wang, Yilun Wang, and Hang Zhao.
\newblock Hdmapnet: An online hd map construction and evaluation framework.
\newblock In \emph{ICRA}, 2022{\natexlab{a}}.

\bibitem[Li et~al.(2022{\natexlab{b}})Li, Wang, Li, Xie, Sima, Lu, Qiao, and Dai]{Zhiqi2022ECCV}
Zhiqi Li, Wenhai Wang, Hongyang Li, Enze Xie, Chonghao Sima, Tong Lu, Yu~Qiao, and Jifeng Dai.
\newblock Bevformer: Learning bird’s-eye-view representation from multi-camera images via spatiotemporal transformers.
\newblock In \emph{ECCV}, 2022{\natexlab{b}}.

\bibitem[Liao et~al.(2023)Liao, Chen, Wang, Cheng, Zhang, Liu, and Huang]{Bencheng2023ICLR}
Bencheng Liao, Shaoyu Chen, Xinggang Wang, Tianheng Cheng, Qian Zhang, Wenyu Liu, and Chang Huang.
\newblock {MapTR}: Structured modeling and learning for online vectorized hd map construction.
\newblock In \emph{ICLR}, 2023.

\bibitem[Liao et~al.(2024)Liao, Chen, Zhang, Jiang, Zhang, Liu, Huang, and Wang]{Bencheng2024IJCV}
Bencheng Liao, Shaoyu Chen, Yunchi Zhang, Bo~Jiang, Qian Zhang, Wenyu Liu, Chang Huang, and Xinggang Wang.
\newblock Maptrv2: An end-to-end framework for online vectorized hd map construction.
\newblock \emph{IJCV}, pages 1--23, 2024.

\bibitem[Liu et~al.(2024{\natexlab{a}})Liu, Wang, Li, Yang, Chen, and Zhu]{Xiaolu2024CVPR}
Xiaolu Liu, Song Wang, Wentong Li, Ruizi Yang, Junbo Chen, and Jianke Zhu.
\newblock Mgmap: Mask-guided learning for online vectorized hd map construction.
\newblock In \emph{CVPR}, 2024{\natexlab{a}}.

\bibitem[Liu et~al.(2023)Liu, Yuan, Wang, Wang, and Zhao]{Yicheng2023ICML}
Yicheng Liu, Tianyuan Yuan, Yue Wang, Yilun Wang, and Hang Zhao.
\newblock Vectormapnet: End-to-end vectorized hd map learning.
\newblock In \emph{ICLR}, 2023.

\bibitem[Liu et~al.(2024{\natexlab{b}})Liu, Zhang, Liu, Zhao, and Xu]{Zihao2024ECCV}
Zihao Liu, Xiaoyu Zhang, Guangwei Liu, Ji~Zhao, and Ningyi Xu.
\newblock Leveraging enhanced queries of point sets for vectorized map construction.
\newblock In \emph{ECCV}, 2024{\natexlab{b}}.

\bibitem[Peng et~al.(2024)Peng, Zhou, Wang, Yang, Chen, and Chen]{Nan2025WACV}
Nan Peng, Xun Zhou, Mingming Wang, Xiaojun Yang, Songming Chen, and Guisong Chen.
\newblock {PrevPredMap}: Exploring temporal modeling with previous predictions for online vectorized hd map construction.
\newblock \emph{arXiv preprint arXiv:2407.17378}, 2024.

\bibitem[Philion and Fidler(2020)]{Jonah2020ECCV}
Jonah Philion and Sanja Fidler.
\newblock Lift, splat, shoot: Encoding images from arbitrary camera rigs by implicitly unprojecting to 3d.
\newblock In \emph{ECCV}, 2020.

\bibitem[Qiao et~al.(2023)Qiao, Ding, Qiu, and Zhang]{Limeng2023CVPR}
Limeng Qiao, Wenjie Ding, Xi~Qiu, and Chi Zhang.
\newblock End-to-end vectorized hd-map construction with piecewise bezier curve.
\newblock In \emph{CVPR}, 2023.

\bibitem[Song et~al.(2024)Song, Chen, Lu, Li, and Skinner]{Jingyu2024ARXIV}
Jingyu Song, Xudong Chen, Liupei Lu, Jie Li, and Katherine~A Skinner.
\newblock {MemFusionMap}: Working memory fusion for online vectorized hd map construction.
\newblock \emph{arXiv preprint arXiv:2409.18737}, 2024.

\bibitem[Wang et~al.(2023)Wang, Liu, Wang, Li, and Zhang]{Shihao2023ICCV}
Shihao Wang, Yingfei Liu, Tiancai Wang, Ying Li, and Xiangyu Zhang.
\newblock Exploring object-centric temporal modeling for efficient multi-view 3d object detection.
\newblock In \emph{ICCV}, 2023.

\bibitem[Wang et~al.(2024)Wang, Jia, Mao, Liu, Zhao, Chen, Wang, Zhang, Zhang, and Zhao]{Shuo2024ECCV}
Shuo Wang, Fan Jia, Weixin Mao, Yingfei Liu, Yucheng Zhao, Zehui Chen, Tiancai Wang, Chi Zhang, Xiangyu Zhang, and Feng Zhao.
\newblock Stream query denoising for vectorized {HD}-map construction.
\newblock In \emph{ECCV}, 2024.

\bibitem[Wilson et~al.(2023)Wilson, Qi, Agarwal, Lambert, Singh, Khandelwal, Pan, Kumar, Hartnett, Pontes, et~al.]{Benjamin2023ARXIV}
Benjamin Wilson, William Qi, Tanmay Agarwal, John Lambert, Jagjeet Singh, Siddhesh Khandelwal, Bowen Pan, Ratnesh Kumar, Andrew Hartnett, Jhony~Kaesemodel Pontes, et~al.
\newblock Argoverse 2: Next generation datasets for self-driving perception and forecasting.
\newblock \emph{arXiv preprint arXiv:2301.00493}, 2023.

\bibitem[Xu et~al.(2024)Xu, K.~Wong, and Zhao]{Zhenhua2024ECCV}
Zhenhua Xu, Kwan-Yee K.~Wong, and Hengshuang Zhao.
\newblock {InsMapper}: Exploring inner-instance information for vectorized hd mapping.
\newblock In \emph{ECCV}, 2024.

\bibitem[Yuan et~al.(2024)Yuan, Liu, Wang, Wang, and Zhao]{Tianyuan2024WACV}
Tianyuan Yuan, Yicheng Liu, Yue Wang, Yilun Wang, and Hang Zhao.
\newblock Streammapnet: Streaming mapping network for vectorized online hd map construction.
\newblock In \emph{WACV}, 2024.

\bibitem[Zhang et~al.(2024)Zhang, Zhang, Ding, Jin, and Yue]{Zhixin2024ECCV}
Zhixin Zhang, Yiyuan Zhang, Xiaohan Ding, Fusheng Jin, and Xiangyu Yue.
\newblock Online vectorized {HD} map construction using geometry.
\newblock In \emph{ECCV}, 2024.

\bibitem[Zhou and Kr{\"a}henb{\"u}hl(2022)]{Brady2022CVPR}
Brady Zhou and Philipp Kr{\"a}henb{\"u}hl.
\newblock Cross-view transformers for real-time map-view semantic segmentation.
\newblock In \emph{CVPR}, 2022.

\bibitem[Zhou et~al.(2024)Zhou, Zhang, Yu, Yang, Jung, Park, and Yoo]{Yi2024CVPR}
Yi~Zhou, Hui Zhang, Jiaqian Yu, Yifan Yang, Sangil Jung, Seung-In Park, and ByungIn Yoo.
\newblock Himap: Hybrid representation learning for end-to-end vectorized hd map construction.
\newblock In \emph{CVPR}, 2024.

\end{thebibliography}
\end{document}